\documentclass[10pt,conference]{IEEEtran}
\usepackage{amsfonts,amsmath, amssymb, graphicx, cite}

\newtheorem{theorem}{Theorem}
\newtheorem{conjecture}{Conjecture}

\title{Limits of Detecting Text Generated by Large-Scale Language Models}
\author{Lav R. Varshney, Nitish Shirish Keskar, and Richard Socher \\ \{lvarshney, nkeskar, rsocher\}@salesforce.com \\
Salesforce Research, Palo Alto, CA, USA}
\date{November 2019}

\begin{document}

\maketitle

\begin{abstract}
Some consider large-scale language models that can generate long and coherent pieces of text as dangerous, since they may be used in misinformation campaigns.  Here we formulate large-scale language model output detection as a hypothesis testing problem to classify text as genuine or generated. We show that error exponents for particular language models are bounded in terms of their perplexity, a standard measure of language generation performance. Under the assumption that human language is stationary and ergodic, the formulation is extended from considering specific language models to considering maximum likelihood language models, among the class of $k$-order Markov approximations; error probabilities are characterized.  Some discussion of incorporating semantic side information is also given.    
\end{abstract}

\section{Introduction}
\label{sec:intro}

Building on a long history of language generation models that are based on statistical knowledge that people have \cite{Shannon1950,Shannon1951,Chapanis1954,JamisonJ1968,TzannesSK1970,CoverK1978}, large-scale, neural network-based language models (LMs) that write paragraph-length text with the coherence of human writing have emerged \cite{RadfordWCLAS2019,KeskarMVXS2019,ZellersHRBFRC2019}.  Such models have raised concerns about misuse in generating fake news, misleading reviews, and hate speech  \cite{SolaimanBCAHWRKKKMNBMW2019, VarshneyKS2019, ZellersHRBFRC2019, BullockL2019, MitchellGSWF2019}.  The alarming consequences of such machine-generated misinformation present an urgent need to discern fake content from genuine, as it is becoming more and more difficult for people to do so without cognitive support tools \cite{GehrmannSR2019}.  Several recent studies have used supervised learning to develop classifiers for this task \cite{ZellersHRBFRC2019,BakhtinGODRS2019,SolaimanBCAHWRKKKMNBMW2019,SchusterSSB2019,IppolitoDCE2019} and interpreted their properties.  Here we take inspiration from our recent work on information-theoretic limits for detecting audiovisual deepfakes generated by GANs \cite{AgarwalV2019} to develop information-theoretic limits for detecting the outputs of language models.  In particular, we build on the information-theoretic study
of authentication \cite{Maurer2000} to use a formal hypothesis testing framework for detecting the outputs  of language models.

In establishing fundamental limits of detection, we consider two settings.  First, we characterize the error exponent for a particular language model in terms of standard performance metrics such as cross-entropy and perplexity.  As far as we know, these informational performance metrics had not previously emerged from a formal operational theorem.
Second, we consider not just a setting with a specific language model with given performance metrics, but rather consider a universal setting where we take a generic view of language models as empirical \emph{maximum likelihood} $k$-order Markov approximations of stationary, ergodic random processes.  Results on estimation of such random processes are revisited in the context of the error probability, using a conjectured extension of the reverse Pinsker inequality.  In closing, we discuss how the semantics of generated text may be a form of side information in detection.

\section{Problem Formulation and Basics}
\subsection{Language Models and their Performance Metrics}
Consider a language $L$ like English, which has tokens drawn from a finite alphabet $\mathcal{A}$; tokens can be letters, words, or other such symbols.  A language model assigns probabilities to sequences of tokens $(a_1,a_2,\ldots,a_m)$ so the more likely a sequence is  in $L$, the greater its probability.  Language models discussed in Sec.~\ref{sec:intro} estimate this probability $Q$ as a product of each token's probability $q$ given its preceding tokens:
\begin{align*}
Q_{(a_1,\ldots,a_n)} &= q(a_1)q(a_2|a_1)q(a_3|a_2,a_1)\cdots q(a_m|a_{m-1},\ldots,a_1) \\
&= \prod_{i=1}^m q(a_i|a_{i-1},\ldots,a_1)\mbox{,}
\end{align*}
a Markov structure.  Some language models including BERT \cite{DevlinCLT2019} consider tokens on both sides in assessing probabilities (cloze task) and are Markov random field models rather than Markov language models \cite{WangC2019}.  Using training data, a language model aims to learn a distribution $Q$ that is close to the empirical distribution $P$ of the language $L$.

Basic language models can be extended to be conditional language models so as to allow control of style.  In certain models, such as CTRL, one can generate text only from a conditioning variable without the need for an initial prompt \cite{KeskarMVXS2019}.  An example of text generated from the conditioning text:

{\tiny  {\tt https://www.cnn.com/2007/09/20/us-president-meets-british-pm} }

\noindent is the following:

{\tiny {\tt  LONDON, England (CNN) -- U.S. President George W. Bush met with British Prime Minister Tony Blair on Monday to discuss the war in Iraq, according to a statement from Blair's office.{\verb \n }{\verb \n }The meeting was held at 10 Downing Street and lasted about an hour.{\verb \n }{\verb \n }"The two leaders discussed Iraq and other international issues of mutual concern," said Blair.{\verb \n }{\verb \n }Bush, who is scheduled to meet Wednesday with Russian President Vladimir Putin, will also visit Germany for talks later this week.{\verb \n }{\verb \n }In his statement, Blair said, "We agreed that we should continue our efforts together to bring peace and stability to Iraq. We both reaffirmed our commitment to working closely together, as well as to continuing to work constructively toward achieving lasting security and prosperity throughout the Middle East region."{\verb \n }{\verb \n }Bush's trip comes after he visited Britain last week where he spoke out against terrorism while visiting Buckingham Palace.{\verb \n }{\verb \n }He has been criticized by some lawmakers over what they say are insufficient military resources being devoted to fighting terrorism.}}

Notwithstanding their limitations \cite{HashimotoZL2019,HoltzmanBFC2019}, the standard performance metrics used for assessing language models are the \emph{cross-entropy} and the \emph{perplexity}, which quantify how close $Q$ is to $P$.  As far as we know, these performance measures have been proposed through the intuitive notion that small values of these quantities seem to correspond, empirically, to higher-quality generated text as judged by people.  Within the common task framework \cite{VarshneyKS2019}, there are leaderboards\footnote{{\tt {\tiny https://paperswithcode.com/sota/language-modelling-on-wikitext-103}}} that assess the perplexity of language models over standard datasets such as WikiText-103 \cite{MerityXBS2016}.  

The cross-entropy of $Q$ with respect to $P$ is defined as:
\[
H(P,Q) = -\mathbb{E}_P[\log Q] \mbox{,}
\]
which simplifies, using standard information-theoretic identities, to:
\[
H(P,Q) = H(P) + D_{\mathrm{KL}}(P || Q) \mbox{,}
\]
where $H(\cdot)$ with one argument is the Shannon entropy and $D_{\mathrm{KL}}( \cdot || \cdot)$ is the Kullback-Leibler divergence (relative entropy).  For a given language $L$ being modeled, the first term $H(P)$ can be thought of as fixed \cite{CoupeODP2019}.  The second term $D_{\mathrm{KL}}(P || Q)$ can be interpreted as the excess information rate needed to represent a language using a mismatched probability distribution \cite{Gilbert1971}. 

Perplexity is also a measure of uncertainty in predicting the next letter and is simply defined as:
\begin{align*}
\mathrm{PPL}(P,Q) &= e^{H(P,Q)} \\
&= e^{H(P)} \cdot e^{D_{\mathrm{KL}}(P || Q)}\mbox{.}
\end{align*}
when entropies are measured in nats, rather than bits.

For a given language, we can consider the ratio of perplexity values or the difference of cross-entropy values of two models $Q_1$ and $Q_2$ as a language-independent notion of performance gap:
\begin{align*}
\mathrm{PPL}(P,Q_1)/\mathrm{PPL}(P,Q_2) &= e^{[D_{\mathrm{KL}}(P || Q_1) - D_{\mathrm{KL}}(P || Q_2)]}\\
&= e^{[H(P,Q_1)-H(P,Q_2)]} \mbox{.}
\end{align*}

\subsection{Hypothesis Test and General Error Bounds}
\label{sec:basics}
Recall that the distribution of authentic text is denoted $P$ and the distribution of text generated by the language model is $Q$.  Suppose we have access to $n$ tokens of generated text from the language model, which we call $Y_1, Y_2, Y_3, \ldots, Y_n$.  We can then formalize a hypothesis test as:
\begin{align*}
H_0 &:= Y \sim P \mbox{ (authentic)} \\
H_1 &:= Y \sim Q \mbox{ (LM generated)}
\end{align*}
If we assume the observed tokens are i.i.d.,  that only makes the hypothesis test easier than the non-i.i.d.\ case seen in realistic text samples, and therefore its performance acts as a bound.

There are general characterizations of error probability of hypothesis tests as follows \cite{CoverT1991}.  For the Neyman-Pearson formulation of fixing the false alarm probability at $\epsilon$ and maximizing the true detection probability, it is known that the error probability satisfies:
\[
\beta^{n}_{\epsilon} \stackrel{.}{=} \exp(-n D_{\mathrm{KL}}(P || Q))
\]
for $n$ i.i.d.\ samples, where $\stackrel{.}{=}$ indicates exponential equality.  Thus the error exponent is just the divergence $D_{\mathrm{KL}}(P || Q))$.  For more general settings (including ergodic settings), the error exponent is given by the asymptotic Kullback-Leibler divergence rate, defined as the almost-sure limit of:
\[
\frac{1}{n} \log \frac{P_n}{Q_n} (y_1, \ldots, y_n)\mbox{, as } n \to \infty\mbox{,}
\]
if the limit exists, where $P_n$ and $Q_n$ are the null and alternate joint densities of $(Y_1,\ldots,Y_n)$, respectively, see further details in \cite{SungTP2006,LuschgyRV1993}.

When considering Bayesian error rather than Neyman-Pearson error, for i.i.d.\ samples, we have the following upper bound: 
\[
P_e^{(n)} \le \exp (-nC(P,Q))
\] 
where $C(\cdot,\cdot)$ is Chernoff information.  Here we will focus on the Neyman-Pearson formulation rather than the Bayesian one.

\section{Limits Theorems}
\label{sec:limits}
With the preparation of Sec.~\ref{sec:basics}, we can now establish statistical limits for detection of LM-generated texts.  We first consider a given language model, and then introduce a generic model of language models.

\subsection{Given Language Model}
Suppose we are given a specific language model such as GPT-2 \cite{RadfordWCLAS2019}, GROVER \cite{ZellersHRBFRC2019}, or CTRL \cite{KeskarMVXS2019}, and it is characterized in terms of estimates of either cross-entropy $H(P,Q)$ or perplexity $\mathrm{PPL}(P,Q)$.

We can see directly that the Neyman-Pearson error of detection in the case of i.i.d.\ tokens is:
\begin{align*}
\beta^{n}_{\epsilon} &\stackrel{.}{=} \exp(-n D_{\mathrm{KL}}(P || Q)) \\
&= \exp(-n (H(P,Q) - H(P))) \mbox{,}
\end{align*}
and similar results hold for ergodic observations.

Since we think of $H(P)$ as a constant, we observe that the error exponent for the decision problem is precisely an affine shift of the cross-entropy.  Outputs from models that are better in the sense of cross-entropy or perplexity are harder to distinguish from authentic text.

Thus we see that intuitive measures of generative text quality match a formal operational measure of indistinguishability that comes from the hypothesis testing limit. 

\subsection{Optimal Language Model}
Now rather than considering a particular language model, we consider bounding the error probability in detection of the outputs of an empirical maximum likelihood (ML) language model.  We specifically consider the empirical ML model among the class of models that are $k$-order Markov approximations of language $L$, which is simply the empirical plug-in estimate.

Manning and Sch\"{u}tze argue that, even though not quite correct, language text can be modeled as stationary, ergodic random processes \cite{ManningS1999}, an assumption that we follow.  Moreover, given the diversity of language production, we assume this stationary ergodic random process with finite alphabet $\mathcal{A}$ denoted $X = \{X_i, -\infty < i < \infty \}$ is \emph{non-null} in the sense that always $P(x_{-m}^{-1}) > 0$ and 
\[
p_m = \inf_{m \ge 1} \min_{a \in \mathcal{A}, x_{-m}^{-1} \in \mathcal{A}^m} P(a|x_{-m}^{-1} ) > 0 \mbox{.}
\]
This is sometimes called the \emph{smoothing requirement}.

We further introduce an additional property of random processes that we assume for language $L$.  We define the \emph{continuity rate} of the process $X$ as:
\begin{align*}
&\gamma(k) \\
&= \sup_{m \ge k} \max_{a\in\mathcal{A}} \max_{x_{-m}^{-1},y_{-m}^{-1} \in\mathcal{A}^m: x_{-m}^{-1}=y_{-m}^{-1}} |P(a|x_{-m}^{-1} ) - P(a|y_{-m}^{-1} )| \mbox{.}
\end{align*}
We further let $\gamma = \sum_{k=1}^{\infty} \gamma(k)$, 
\[
\alpha = \frac{1}{\prod_{j=1}^{\infty}} (1 - \gamma(j)) \mbox{,}
\]
and
\[
\beta(k) = \frac{1 - (1-|\mathcal{A}|\gamma(k))^k}{k\gamma(k) \prod_{j=1}^{\infty} (1-|\mathcal{A}|\gamma(j))^2} \mbox{.}
\]
If $\gamma < \infty$, then the process has \emph{summable continuity rate}.  These specific technical notions of smoothing and continuity are taken from the literature on estimation of stationary, ergodic random processes \cite{CsiszarT2010}.

As such, the hypothesis test we aim to consider here is between a non-null, stationary, ergodic process with summable continuity rate (genuine language) and its \emph{empirical} $k$-order Markov approximation based on training data (language model output).  We think of the setting where the language model is trained on data with many tokens, a sequence of very long length $m$.  For example, the CTRL language model was trained using 140 GB of text \cite{KeskarMVXS2019}.

We think of the Markov order $k$ as a large value and so the family of empirical $k$-order Markov approximations encompasses the class of neural language models like GPT-2 and CTRL, which are \emph{a fortiori} Markov in structure.  Empirical perplexity comparisons show that LSTM and similar neural language models have Markov order as small as $k = 13$ \cite{ChelbaNB2017}.  The appropriate Markov order for large-scale neural language models has not been investigated empirically, but is thought to scale with the neural network size.  

Now we aim to bound the error exponent in hypothesis testing, by first drawing on a  bound for the Ornstein $\bar{d}$-distance between a stationary, ergodic process and its Markov approximation, due to Csiszar and Talata \cite{CsiszarT2010}.  Then we aim to relate the Ornstein $\bar{d}$-distance to the Kullback-Leibler divergence (from error exponent expressions), using a generalization of the so-called reverse Pinsker inequality \cite{SasonV2016,Binette2019}. 

Before proceeding, let us formalize a few  measures.  Let the per-letter Hamming distance between two strings $x_1^m$ and $y_1^m$ be $d_m(x_1^m,y_1^m)$.  Then the Ornstein $\bar{d}$-distance between two random sequences $X_1^m$ and $Y_1^m$ with distributions $P_X$ and $P_Y$ is defined as:
\[
\bar{d}(X_1^m,Y_1^m) = \min_{\mathbb{P}} \mathbb{E}_{\mathbb{P}} d_m(\tilde{X}_1^m,\tilde{Y}_1^m) \mbox{,}
\]
where the minimization is over all joint distributions whose marginals equal $P_X$ and $P_Y$.

Let $N_m(a_1^k)$ be the number of occurrences of the string $a_1^k$ in the sample $X_1^m$.  Then the empirical $k$-order Markov approximation of a random process $X$ based on the sample $X_1^m$ is the stationary Markov chain of order $k$ whose transition probabilities are the following empirical conditional probabilities:
\[
\hat{P}_m(a|a_1^k) = \frac{N_m(a_1^k a)}{ N_{m-1}(a_1^k)} \mbox{, } a \in \mathcal{A} \mbox{ and } a_1^k \in \mathcal{A}^k \mbox{.}
\]
We refer to this empirical approximation as $\hat{X}[k]_1^m$.

Although they give more refined finitary versions, let us restate Csisz{\'{a}}r and Talata's asymptotic result on estimating Markov approximations of stationary, ergodic processes from data.  The asymptotics are in the size of the training set, $m \to \infty$, and we let the Markov order scale logarithmically with $m$.
\begin{theorem}[\cite{CsiszarT2010}]
Let $X$ be a non-null stationary ergodic process with summable continuity rate.  Then for any $\nu > 0$, the empirical $(\nu \log m)$-order Markov approximation $\hat{X}$ satisfies:
\[
\bar{d}(X_1^m,\hat{X}[\nu \log m]_1^m) \le \frac{\beta (\nu \log m)}{p_m^2} \gamma (\nu \log m) + \frac{1}{m^{1/2 - \mu}}
\]
eventually almost surely as $m\to\infty$ if $\nu < \tfrac{\mu}{|\log p_m|}$.
\end{theorem}


Now we consider Kullback-Leibler divergence.  Just as Marton had extended Pinsker's inequality between variational distance and Kullback-Leibler divergence to an inequality between Ornstein's $\bar{d}$-distance and Kullback-Leibler divergence \cite{Marton1996, Marton1998} as given in Theorem~\ref{thm:marton} below, is it possible to make a similar conversion for the reverse Pinsker inequality when there is a common finite alphabet $\mathcal{A}$?  
\begin{theorem}[\cite{Marton1998}]
\label{thm:marton}
Let $X$ be a stationary random process from a discrete alphabet $\mathcal{A}$. Then for any other random process $Y$ defined on the same alphabet $\mathcal{A}$, 
\[
\bar{d}(X_1^m,Y_1^m) \le (u + 1) \sqrt{\tfrac{1}{2m} D(X_1^m \| Y_1^m)} 
\]
for a computable constant $u$.
\end{theorem}

We conjecture that one can indeed convert the reverse Pinsker inequality \cite{SasonV2016}:
\[
D(P \| Q) \le \frac{\log e}{Q_{\min}} |P - Q|^2
\]
for two probability distributions $P$ and $Q$ defined on a common finite alphabet $\mathcal{A}$, 
where $Q_{\min} = \min_{a\in\mathcal{A}} Q(a)$.  That is, we make the following conjecture.
\begin{conjecture}
Let $X$ be a stationary random process from a finite alphabet $\mathcal{A}$. Then for any other random process $Y$ defined on the same alphabet $\mathcal{A}$, 
\[
D(X_1^m \| Y_1^m) \le \tilde{K} \ \bar{d}(X_1^m , Y_1^m)^2
\]
for some constant $\tilde{K}$.
\end{conjecture}

If this generalized reverse Pinsker inequality  holds, it implies the following further bound on the Kullback-Leibler divergence and therefore the error exponent of the detection problem for the empirical maximum likelihood Markov language model.
\begin{conjecture}
Let $X$ be a non-null stationary ergodic process with summable continuity rate defined on the finite alphabet $\mathcal{A}$. Then for any $\nu > 0$, the empirical $(\nu \log m)$-order Markov approximation $\hat{X}$ satisfies:
\begin{align*}
&D(X_1^m \| \hat{X}[\nu \log m]_1^m) \\
&\quad\le \hat{K} \left\{\frac{\beta (\nu \log m)}{p_m^2} \gamma (\nu \log m) + \frac{1}{m^{1/2 - \mu}} \right\}^2
\end{align*}
eventually almost surely as $m\to\infty$ if $\nu < \tfrac{\mu}{|\log p_m|}$, for some constant $\hat{K}$.
\end{conjecture}
Under the conjecture, we have a precise asymptotic characterization of the error exponent in
deciding between genuine text and text generated from the empirical maximum likelihood language model, expressed in terms of basic parameters of the language, and of the training data set.

\section{Discussion}
Motivated by the problem of detecting machine-generated misinformation text that may have deleterious societal consequences, we have developed a formal hypothesis testing framework and established limits on the error exponents.  For the case of specific language models such as GPT-2 or CTRL, we provide a precise operational interpretation for the perplexity and cross-entropy.  For any future large-scale language model, we also conjecture a precise upper bound on the error exponent.  

It has been said that ``in AI circles, identifying fake media has long received less attention, funding and institutional backing than creating it: Why sniff out other people’s fantasy creations when you can design your own? `There's no money to be made out of detecting these things,' [Nasir] Memon said'' \cite{Harwell2019}.  Here we have tried to demonstrate that there are, at least, interesting research questions on the detection side, which may also inform practice.

As we had considered previously in the context of deepfake images \cite{AgarwalV2019}, it is also of interest to understand how error probability in detection parameterizes the dynamics of information spreading processes in social networks, e.g.\ in determining epidemic thresholds.

Many practical fake news detection algorithms use a kind of \emph{semantic} side information, such as whether the generated text is factually correct, in addition to its statistical properties.  Although statistical side information would be straightforward to incorporate in the hypothesis testing framework, it remains to understand how to cast such semantic knowledge in a statistical decision theory framework.

\section*{Acknowledgment}
Discussions with Bryan McCann, Kathy Baxter, and Miles Brundage are appreciated.

\bibliographystyle{IEEEtran}
\bibliography{ethics_socialgood,detection}

\end{document}